\documentclass[11pt]{article}
\usepackage{euler}
\usepackage{euscript}
\usepackage{fullpage}
\usepackage{euler}
\usepackage{color}
\usepackage{verbatim}

\usepackage{amsmath}
\usepackage{amsfonts}
\usepackage{amssymb}
\usepackage{xspace}
\usepackage{graphicx} 
\usepackage{wrapfig} 
\usepackage[numbers]{natbib}
\usepackage{algorithm}
\usepackage[noend]{algorithmic}
\usepackage{hyperref}
\usepackage{xspace}
\usepackage{euscript}
\usepackage{paralist}
\usepackage{color,xcolor}
\usepackage{subfigure}
\usepackage[labelfont=sf]{caption}

\newtheorem{defn}{Defn}[section]

\usepackage[draft,inline,nomargin,index]{fixme}
\fxsetup{theme=colorsig,mode=multiuser,inlineface=\sffamily,envface=\itshape}
\FXRegisterAuthor{sv}{asv}{Suresh}
\FXRegisterAuthor{pr}{apr}{Parasaran}

\newcommand{\etal}{et al.\xspace}
\newcommand{\affinityalgo}{\textsc{Affinity}\xspace}

\renewcommand{\b}[1]{\ensuremath{\mathbb{#1}}}

\newcommand{\reals}{\b{R}}
\newcommand{\vol}{\ensuremath{\text{Vol}}}
\newcommand{\SamplePolytope}{\textbf{SamplePolytope}}
\newcommand{\hitandrun}{\textbf{Hit-And-Run}\xspace}

\newcommand{\eps}{\varepsilon}

\newcommand{\liftemd}{\textsc{LiftEMD}\xspace}

\begin{document}

\title{Power to the Points: \\
Validating Data Memberships in Clusterings}

\author{
Parasaran Raman\\
       {School of Computing}\\
       {University of Utah}\\
       {praman@cs.utah.edu}
\and 
Suresh Venkatasubramanian\\
       {School of Computing}\\
       {University of Utah}\\
       {suresh@cs.utah.edu}     
}

\maketitle

\begin{abstract}
A clustering is an implicit assignment of labels of points, based on proximity to other points. It is these labels that are then used for downstream analysis (either focusing on individual clusters, or identifying representatives of clusters and so on). Thus, in order to trust a clustering as a first step in exploratory data analysis, we must trust the labels assigned to individual data. Without supervision, how can we validate this assignment? 

In this paper, we present a method to attach affinity scores to the implicit labels of individual points in a clustering. The affinity scores capture the confidence level of the cluster that claims to "own" the point. This method is very general: it can be used with clusterings derived from Euclidean data, kernelized data, or even data derived from information spaces. It smoothly incorporates importance functions on clusters, allowing us to weight different clusters differently. It is also efficient: assigning an affinity score to a point depends only polynomially on the number of clusters and is independent of the number of points in the data. The dimensionality of the underlying space only appears in preprocessing. 

We demonstrate the value of our approach with an experimental study that illustrates the use of these scores in different data analysis tasks, as well as the efficiency and flexibility of the method. We also demonstrate useful visualizations of these scores; these might prove useful within an interactive analytics framework.

\end{abstract}

\section{Introduction}
\label{sec:intro}

We live in an era of personalized data mining, where the power and sophistication of learning tools are used to make predictions for individuals. Whether it be Netflix or Amazon recommendations, customized medical diagnosis based on genetic markers, or even decisions on whether or not place someone on a no-fly list, important decisions are being made at the \emph{individual} level based on techniques that have \emph{global} predictive guarantees. These decisions impact our lives in ways large and small, and their impact is greater because they are personalized. It is therefore very important that we have \emph{personalized} ways to validate these decisions. 

Clustering is an unsupervised exploratory data mining technique that generates predictions in the form of implicit labels for points. These predictions are used for exploration, data compression, and further analysis, and so it is important to verify the accuracy of the labels. Clustering is unsupervised however, and there is no direct way to validate the data assignments. Thus, a number of indirect approaches have been developed to validate a clustering at a \emph{global} level\cite{Xu:2009:CLU:1483087}. These include internal, external and relative validation techniques, and methods based on \emph{clustering stability} that assume a clustering (algorithm) is good if small perturbations in the input do not affect the output clustering significantly.

But all these approaches are global: they assign a single number to a clustering. While this perspective might be suitable from the perspective of the entity running the algorithm, it is insufficient to answer the question: is \emph{my} data correctly assigned ? 

\textbf{Example.} Consider a service like Klout that tracks user reputation on social media and offers perks tailored to users in particular categories. For example, a blogger who generates a lot of traffic for their posts on home repair might be targeted with a discount code for Home Depot. In such a setting, knowing that  the overall grouping of users into different occupation categories is accurate does not help in suggesting recommendations. What Klout wants is a sense of how ``close'' an individual is to a particular grouping, so that the targeting is more effective. 

\textbf{Example.} Consider a less benign example. It is now possible to take large amounts of genetic data and build screens that identify people with risk factors for various diseases. If your data falls into a high risk cluster, you might potentially be denied access to insurance, or forced to pay a premium. In such a setting, it is important to know at a \emph{local} level if the assignment being made is valid or not. 

While these examples are a little stylized, they illustrate an important point. Whether we use clustering as a exploratory tool to focus attention, or as a ``last-step'' prediction tool, local validation differs greatly from global validation. 

\paragraph{Desiderata}
\label{sec:desiderata}

A measure of local validity is a number we assign to an individual point of a clustering. What properties should such a validation provide ? It should have a well-defined scale. It will incorporate both spatial and combinatorial aspects of the clustering (like distances between clusterings). We should be able to define it for a single clustering of any kind without generative assumptions, but we should also be able to compare scores for a point if we have different candidate clusterings. Examples of such measures exist for other prediction tasks. For example, the margin of an individual point in binary classification (the distance from the point to the decision boundary) is a measure that satisfies some of the above properties\footnote{While it is not scale-invariant, one can scale the data by fixing the classifier margin to be $1$ (so that all points have a margin of at least $1$).}. For clustering, the problem is more challenging because there are multiple decision boundaries (with respect to each cluster) and these need to be combined in a meaningful way. 

\textbf{Validation versus outlier detection.} Local validation bears a superficial resemblance to outlier detection: in both cases the goal is to evaluate individual points based on how well they ``fit'' into a clustering. There are important differences though. An outlier affects the \emph{cost} of a clustering by being far away from any cluster, but it will usually be clear what cluster it might be assigned to. In contrast, a point whose \emph{labelling} might be invalid is usually in the midst of the data. Assigning it to one cluster or another might not actually change the clustering cost, even though the label itself is now unreliable.

\subsection{Our Work}
\label{sec:our-work}

We now introduce the key ideas underlying our proposed method for determining local validity. The key notion in clustering is \emph{proximity}: points are expected to have similar labels if they are close to each other and not to others. In other words, the \emph{regions of influence} of  points belonging to the same cluster must overlap. \cite{houle-ssdbm} 

Therefore, a point should be associated with a cluster if its region of influence significantly overlaps the region of influence of the cluster, and does not have such an overlap with other clusters. And more importantly, we can quantify the confidence of this association by measuring the \emph{degree} of overlap. 

The way we estimate this quantity is via a thought experiment: suppose the point under consideration was a separate cluster by itself ? How  much of its influence would be derived from the set of existing clusters ? Answering this question will lead to the measure we propose. 

Consider for example using the distance to a cluster center as a measure of influence. Then a point is most influenced by clusters that it is closest to. The distance here acts as a one-dimensional measure of influence: if all points are on the line, then the region of influence of a point can be quantified as half the distance to the nearest point. 

The method we propose is a generalization of this idea to incorporate a variety of more general notions of regions of influence that can incorporate cluster importance, density and even different cluster shapes. The key idea is to define regions of influence as elements of an appropriate weighted power diagram (a generalization of a Voronoi diagram) and use random sampling to estimate how regions of influence overlap (and avoid the curse of dimensionality). 
\subsection{Applications}
\label{sec:applications}

Our belief is that the ability to quantify the validity of a clustering locally, in addition to the above mentioned applications, will aid in existing tools for clustering and metaclustering. We present a few examples. 

\textbf{Active Clustering.} Active learning\cite{active-survey,hofmann1998active,eriksson2011active} is a supervised learning technique that assumes that labels are hard to obtain, and so tries to learn a task while querying as few labels as possible. While clustering is unsupervised, a variant of this idea applies here as well. In this case, since clustering is expensive, the goal is select points that will be most influential in deciding the final decision boundaries of the clustering. Our method can be used to select points that are least affiliated with any current clustering.

\textbf{Validating Consensus.} A \emph{consensus clustering}\cite{Strehl:2003:CEK:944919.944935} is the consensus answer obtained from a collection of clusterings (arising from different runs of an algorithm or different algorithms). A consensus clustering is only useful if the underlying clusterings mostly agree with each other, else the resulting clustering might have no connection to the original clusterings. But there's no easy way to test this ! Our procedure can be used to validate consensus clusterings by verifying that the number of valid points increases when we do consensus.

\textbf{Incremental Clustering.} Many clustering algorithms for large data proceed incrementally\cite{dbscan,birch,DBLP:reference/db/Venkatasubramanian09}. A partial clustering is built up from the data seen thus far, and then the new data updates this clustering. By identifying points with weak validity, we can treat them differently to points that are strongly associated with clusters by maintaining them separately without committing them to a cluster. Points strongly associated with a cluster can then be collapsed to the cluster centers safely.

\section{Background}

There have been a number of approaches to testing \emph{global} validity of a clustering. These can roughly be broken down into three categories\cite{Xu:2009:CLU:1483087}. \emph{Internal validation mechanisms} look at the structure of a clustering and attempt to determine its quality. For example, the ratio of the minimum inter-cluster distance to the maximum intra-cluster distance is a measure of how well-separated clusters are, and thus how good the clustering is.\emph{External} validation measures can be employed when a reference clustering exists. In this case, an appropriate distance between clusterings must be defined, and then the given clustering can be compared to the reference clustering. \emph{Relative} validation measures look at different runs of a clustering algorithm and compare the resulting clusterings produced. 

Another approach to understanding clustering quality is via the idea of \emph{cluster stability}\cite{DBLP:conf/colt/Ben-DavidLP06,DBLP:conf/psb/Ben-HurEG02,group:2572:Bezdek1998}. The goal here is to determine how robust a clustering solution is to small perturbations in algorithm parameters. This idea was used to do model selection; for example, the ``right'' number of clusters is the one that exhibits the most stable clusterings. 

Stability in general has been studied extensively in the statistics and machine learning communities, as a way to understand generalization properties of algorithms. The paper by Elisseeff \etal\cite{elisseeff2006stability} provides a good overview of this literature and the monograph by Luxburg\cite{von2010clustering} focuses on clustering.

section{Preliminaries}
\label{sec:preliminaries}

Let $P$ be a set of $n$ points in $\reals^d$. We assume a distance measure $D$ on $\reals^d$, which for now we will take to be the Euclidean distance. A \emph{clustering} is a partition of $P$ into clusters $C_1, C_2, \ldots, C_k$. We will assume that we can associate a representative $c_i$ with a cluster $C_i$. For example, the representative could be the cluster centroid, or the median. We will denote the set of representatives by $C$.

A \emph{Voronoi diagram} on a set of sites $S = \{s_1, s_2, \ldots, s_k\} \subset \reals^d$ is a partition of $\reals^d$ into regions $V_1, \ldots V_{k}$ such that for all points in $V_i$, the site $s_i$ is the closest neighbor. Formally, $V_i = \{ p \in \reals^d | D(p, s_i) \le D(p, s_j), j \ne i\}$. When $D$ is the Euclidean distance, the boundary between two regions is always a hyperplane, and therefore each cell $V_i$ is a convex polyhedron with at most $k-1$ faces. 

\begin{figure}
  \centering
  \includegraphics[width=4in]{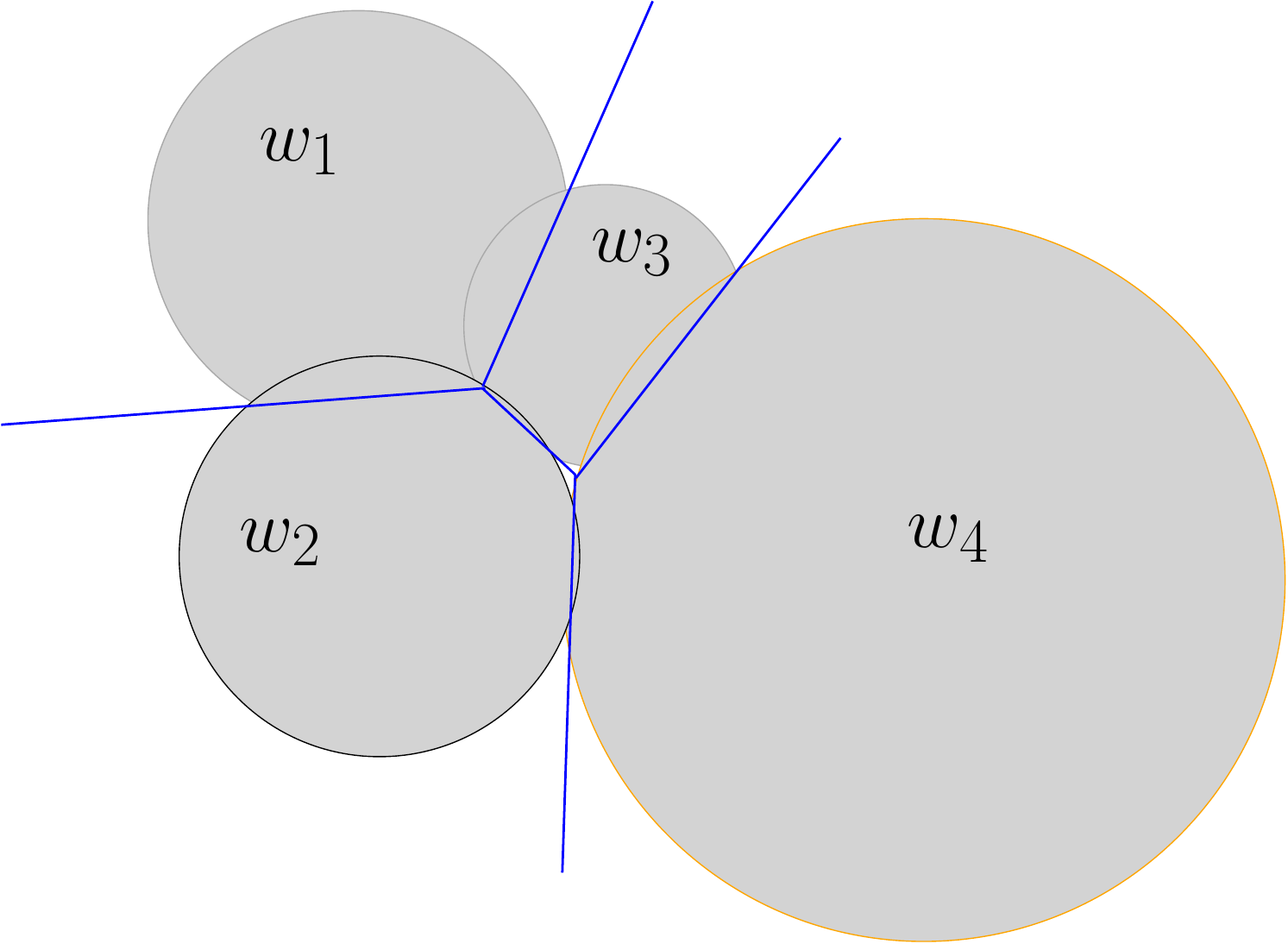}
  \caption{The power diagram of a set of points. The sphere radius is proportional to the weight $w$}
  \label{fig:powerdiag}
\end{figure}

We will also make use of a generalization of the Voronoi diagram called the \emph{power diagram}. Suppose that we associate an importance score $w_i$ with each site $s_i$. Then the power diagram on $S$ (see Figure~\ref{fig:powerdiag}) is also a partition of $\reals_d$ into $k$ regions $V_i$, such that $V_i = \{ p \in \reals^d \mid D^2(p, s_i) - w_i \le D^2(p, s_j)  -w_j, j \ne i\}$. 

Power diagrams allow us to give different sites different influence, but retain the property that all boundaries between regions are hyperplanes and all regions are polyhedra in Euclidean space\footnote{The squared distance is crucial to making this happen; without it, arcs could be elliptical or hyperbolic.}.

Finally, we will frequently refer to the \emph{volume} $\vol(S)$ of a region $S \subset \reals^d$. In general, this denotes the $d$-dimensional volume of $S$ with respect to the standard Lebesgue measure on $\reals^d$. If $S$ is not full-dimensional, this should be understood as referring to the lower-dimensional volume, or the volume of the relative interior of $S$; for example the ``volume'' of a triangle in three dimensions is its area, and the volume of a line segment is its length, and so on. 

\section{Defining Affinity Scores}

As we discussed in Section~\ref{sec:intro}, the \emph{region of influence} of a point is how we define its affinity to clusters. Each cluster has a region of influence. If we now consider a particular point in the data and treat it as a singleton cluster, it will draw influence from neighboring clusters (for example, in a $k$-means setting, points assigned to other clusters might be reassigned to it). If a point draws its influence primarily from one cluster, then its affinity with that cluster should be high. More generally, the affinity of a point to a cluster will derive from the \emph{proportion of influence it steals from that cluster. }

\begin{defn}[Region of Influence]
  Let $\mathcal{C} = C_1, C_2, \ldots C_k$ be a clustering of $n$ points. A \emph{region of influence function} is a function $R : \mathcal{C} \rightarrow  2^{\reals^d}$ on $\mathcal{C}$ such that all $R(C_i)$ (which are subsets of $\reals^d$) are disjoint.
\end{defn}
\begin{defn}[Affinity Scores]
Let $R$ be a region-of-influence function. Let $\mathcal{C} = C_1, C_2, \ldots C_k$ be a clustering. For any point $x$, let $\mathcal{C}_x$ denote the clustering $C_1 \setminus \{x\}, C_2 \setminus \{x\}, \ldots, C_k \setminus \{x\}, \{x\}$, and let $R_x(C)$ denote the region of influence of a cluster $C \in \mathcal{C}_x$. 

Then the affinity score of $x$ is the vector $(\alpha_1, \alpha_2, \ldots, \alpha_k)$, where 
\[ \alpha_i = \frac{\vol(R(C_i) \cap R_x(\{x\}))}{\vol(R_x(\{x\}))} \]
\end{defn}
In the above definition, $R_x(\{x\})$ is the region of influence $x$ has carved out for itself, and $\alpha_i$ merely captures the proportion of $R_x(\{x\})$ that comes from the (original) cluster $C_i$. 

There can be at most one $\alpha_i > 1/2$. If such an $i$ exists, we say that $x$ is \emph{stable}: intuitively, cluster $C_i$ wins the vote over other clusters to own $x$. If not, we refer to $\max_i \alpha_i$ as the \emph{affinity score} of $x$, and set the affinity score of a stable point to $1$. The lower the affinity score, the less confident we are about the label for $x$. 

The simplest region of influence function is a Voronoi cell. Specifically, consider a clustering with $k$ clusters, each cluster $C_i$ having representative $c_i$. Let $C$ be the set of these representatives. 
Consider any point $x \in CH(C)$ (the convex hull of $C$). Let $V_1, V_2, \ldots, V_k$ be the Voronoi partition of $C$, and let $V'_1, V'_2, \ldots, V'_k, V'_x$ be the Voronoi partition of $C \cup \{x \}$, with $V'_x$ being the Voronoi cell of $x$. Then we define $R(C_i) = V_i$, and $R_x(C_i) = V'_i$. We call a hyperplane supporting $V_x$ that separates $c_i$ from $x$ a \emph{supporting hyperplane} for $x$ with respect to $c_i$.

\begin{figure}[htbp]
  \centering
  \includegraphics[width=4in]{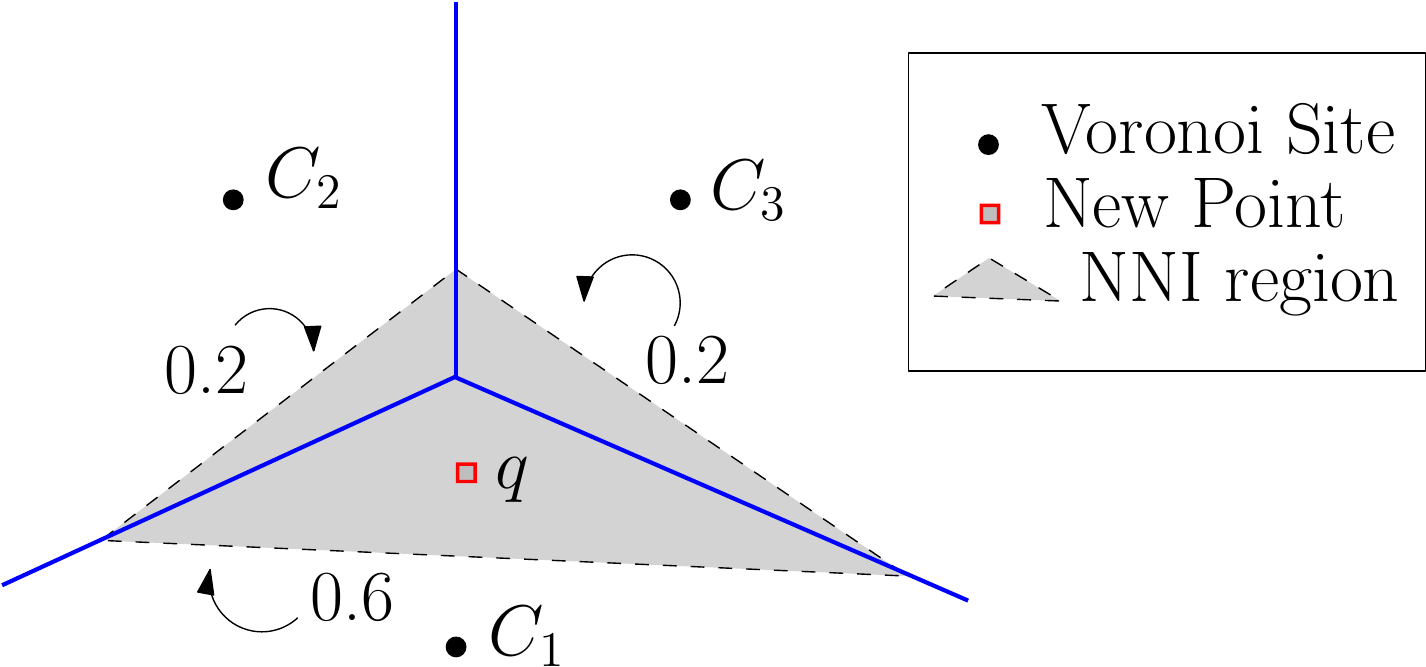}
  \caption{In this example, the red point is ``stealing'' the shaded area from the Voronoi cells of $C_1, C_2, C_3$. }
  \label{fig:areastealing}
\end{figure}
The Voronoi cell $V'_x$ of $x$ ``steals'' volume from Voronoi cells around it (Figure~\ref{fig:areastealing} illustrates this concept). We can compute the fraction of $V'_x$ that comes from any other cell. For any point $p_i \in P$, let $\alpha_i = \frac{\vol(V_i \cap V'_x)}{\vol(V_x)}$. Then $\alpha_i$ represents the (relative) amount of volume that $x$ ``stole'' from $p_i$. Note that $\sum \alpha_i = 1$, and if $x = p_i$, then $\alpha_i = 1$. 

\paragraph{Note} This idea of \emph{area stealing} was first defined in the context of natural neighbor interpolation\cite{nni1,nni2}, where the $\alpha_i$ values were then used to compute an interpolation of function values at the $p_i$. However, in this paper we will use the $\alpha_i$ directly, without computing any interpolants.

\subsection{Affinity versus Distance}

\begin{figure}[htbp]
  \centering
  \includegraphics[width=4in]{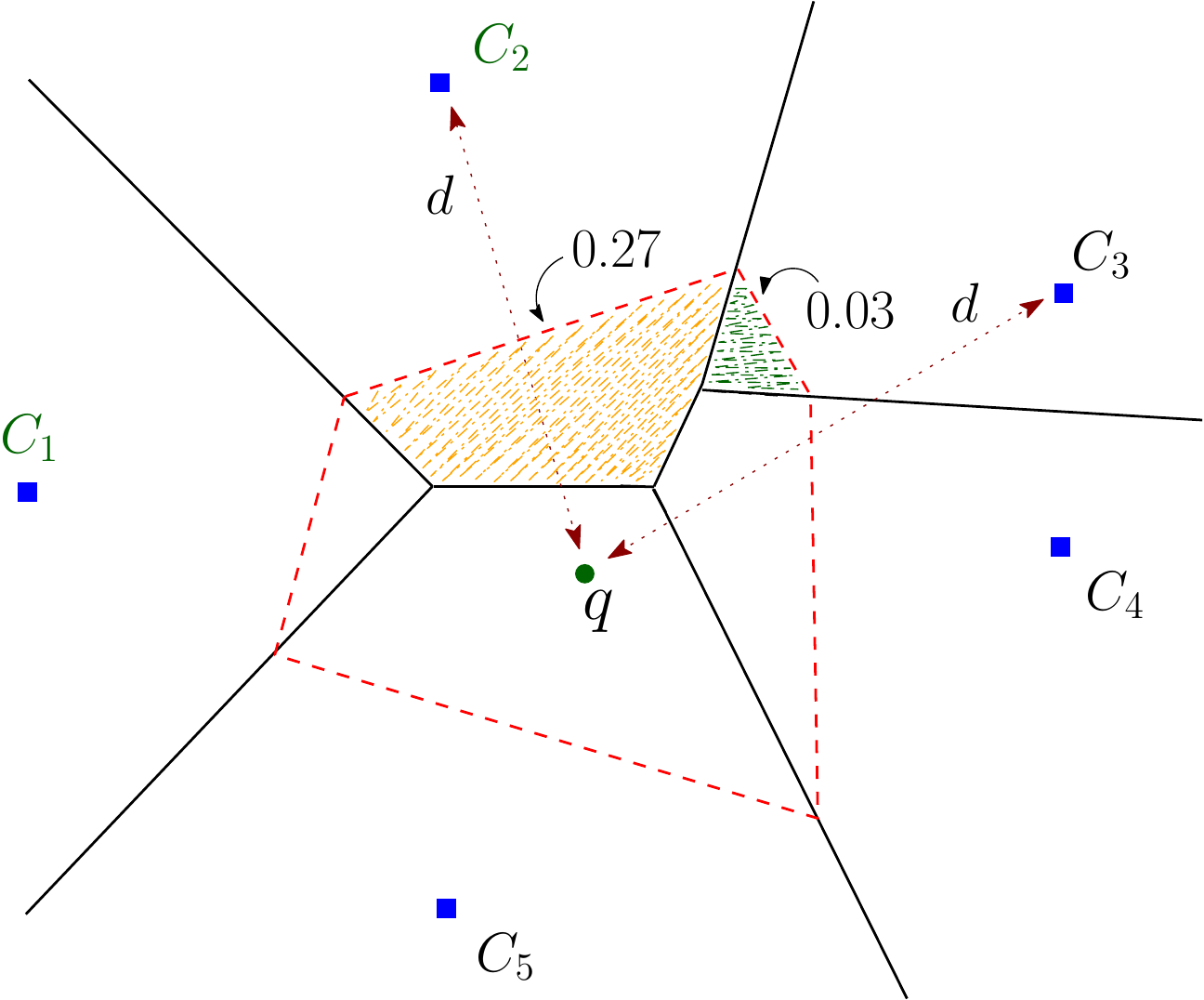}
  \caption{Illustration of the difference between distance-based and area-based influence measures}
  \label{fig:spatial}
\end{figure}

The simplest way to define influence is by distance. For example, we could define the affinity of a point to a cluster as the (normalized) distance between the point and the cluster representative. We note that affinity generalizes distance ratios: in one dimension, affinity calculations yield the same result as distance ratios, since the ``area'' stolen from a cell is merely half the distance to that cell. But distances cannot capture influence created by the spatial relationship of the clusters. Consider the configuration shown in Figure~\ref{fig:spatial}. The point $q$ is equidistant from the cluster centers $c_2$ and $c_3$ and so would have the same distance-based influence with respect to these clusters. But when we examine the configuration more closely, we see that the presence of $c_4$ is reducing the influence of $c_3$ on $q$, and this effect appears only when we look at a \emph{planar} region of influence. We validate this by 100 runs of $k$-means with random seeds and watch which clusters $q$ ends up in. We observe that $q$ was assigned to $c_2$ in 15 runs and to $c_3$ in only 2 runs. Note that heuristically, the distance ratios behave like the volume ratios to the power of $1/d$, and so the distance ratios would be expected to show less variation (and therefore less fidelity) across clusters. 

\subsection{Visualization}
\label{sec:visualization}

\begin{figure}[htbp]
  \centering
  \includegraphics[width=4in]{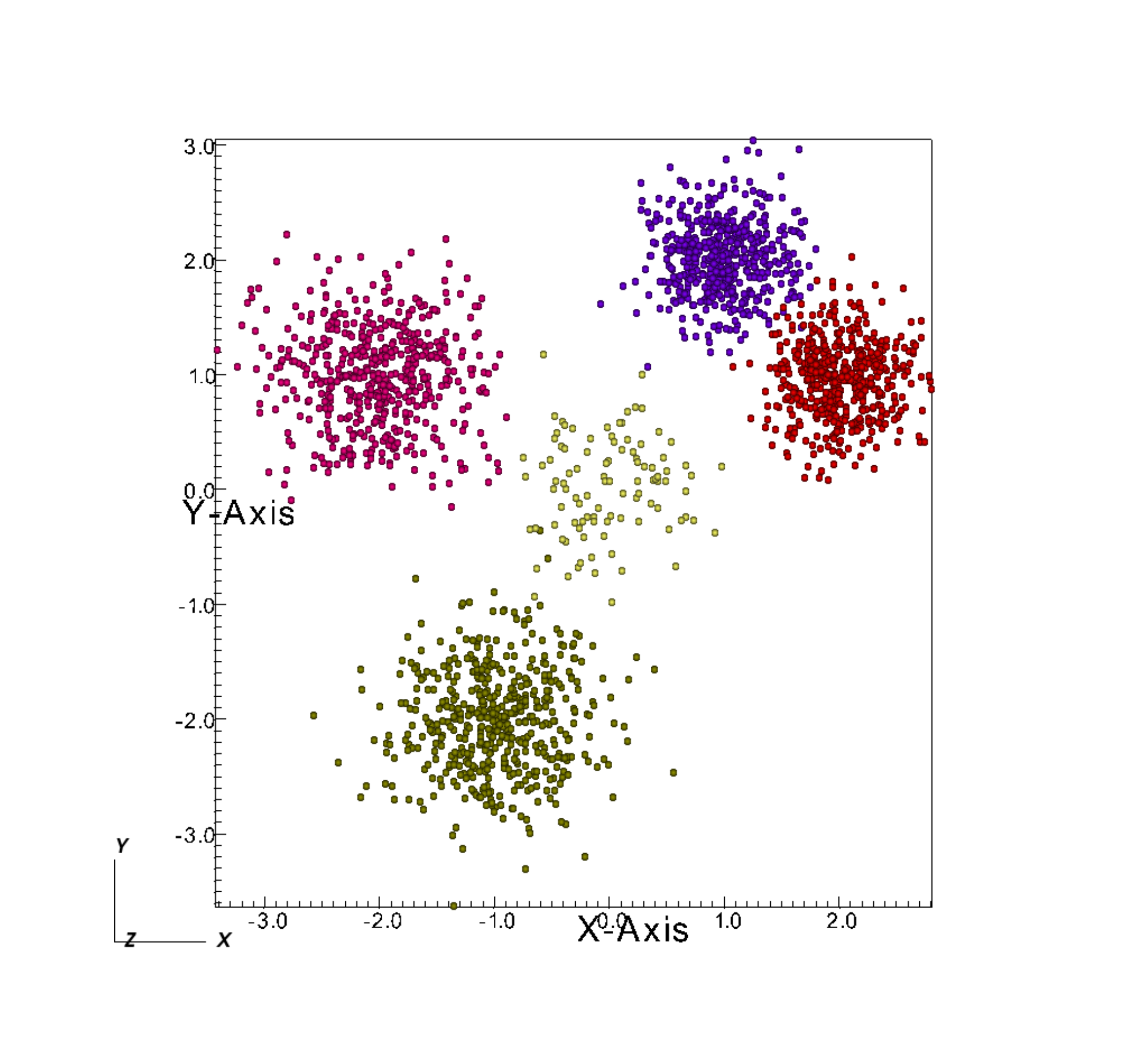}
  \caption{Data in five clusters}
  \label{fig:viz1}
\end{figure}

The affinity scores define a scalar field over the space the data is drawn from, and can be visualized (in low dimensions). Consider the clustering depicted in Figure\ref{fig:viz1}. We can draw a contour map where each level connects points with the same affinity score (unlike in a topographical map, more deeply nested contours correspond to \emph{lower} affinity scores). We can also render this as a greyscale heatmap (where the lower the affinity, the brighter the color). These visualizations, while simple, provide a nice visual rendering of affinity scores that can be useful as part of the exploratory analysis pipeline.

\begin{figure}[htbp]
  \centering
  \includegraphics[width=4in]{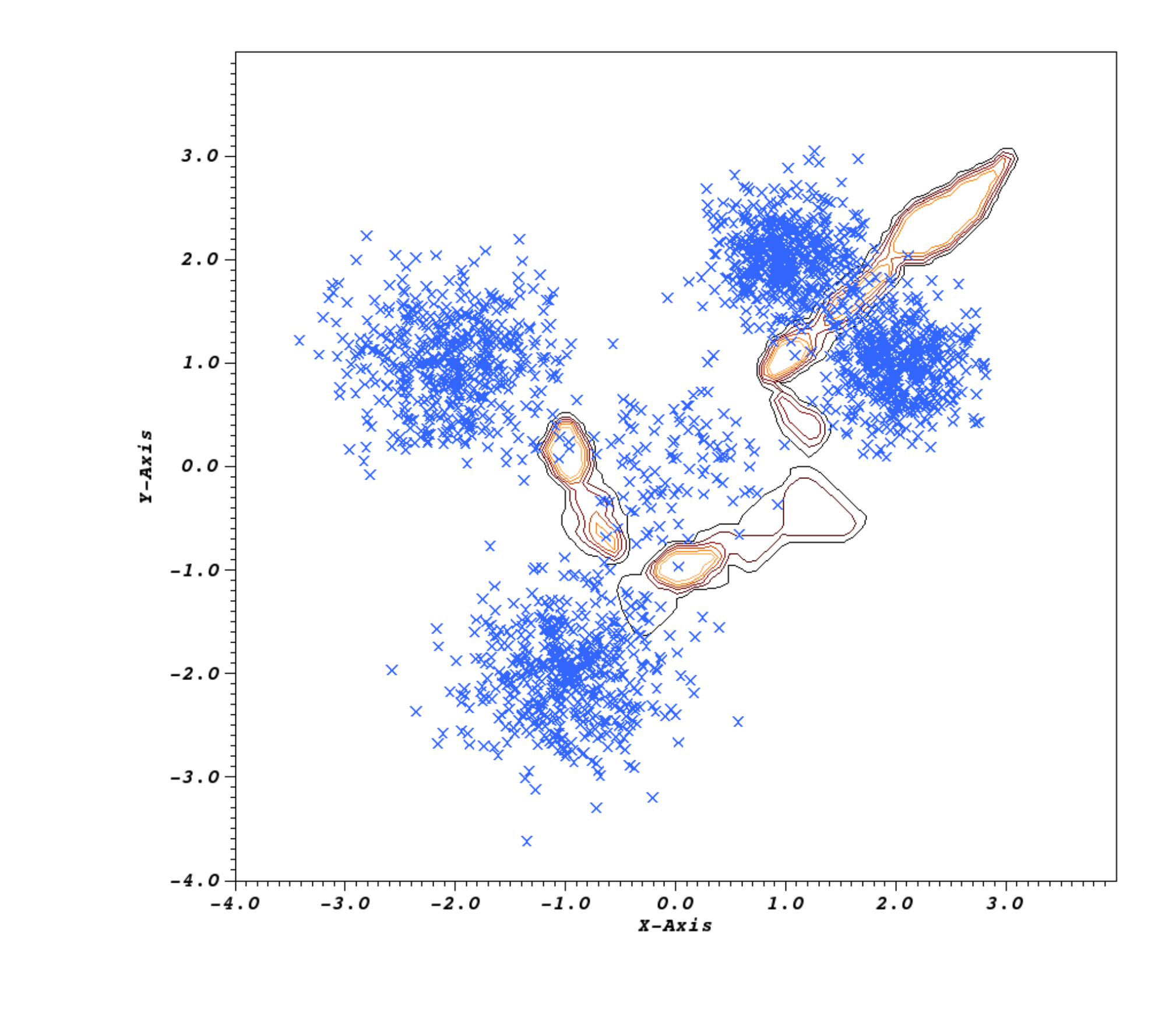}
  \caption{A contour plot}
  \label{fig:viz2}
\end{figure}

\begin{figure}[htbp]
  \centering
  \includegraphics[width=4in]{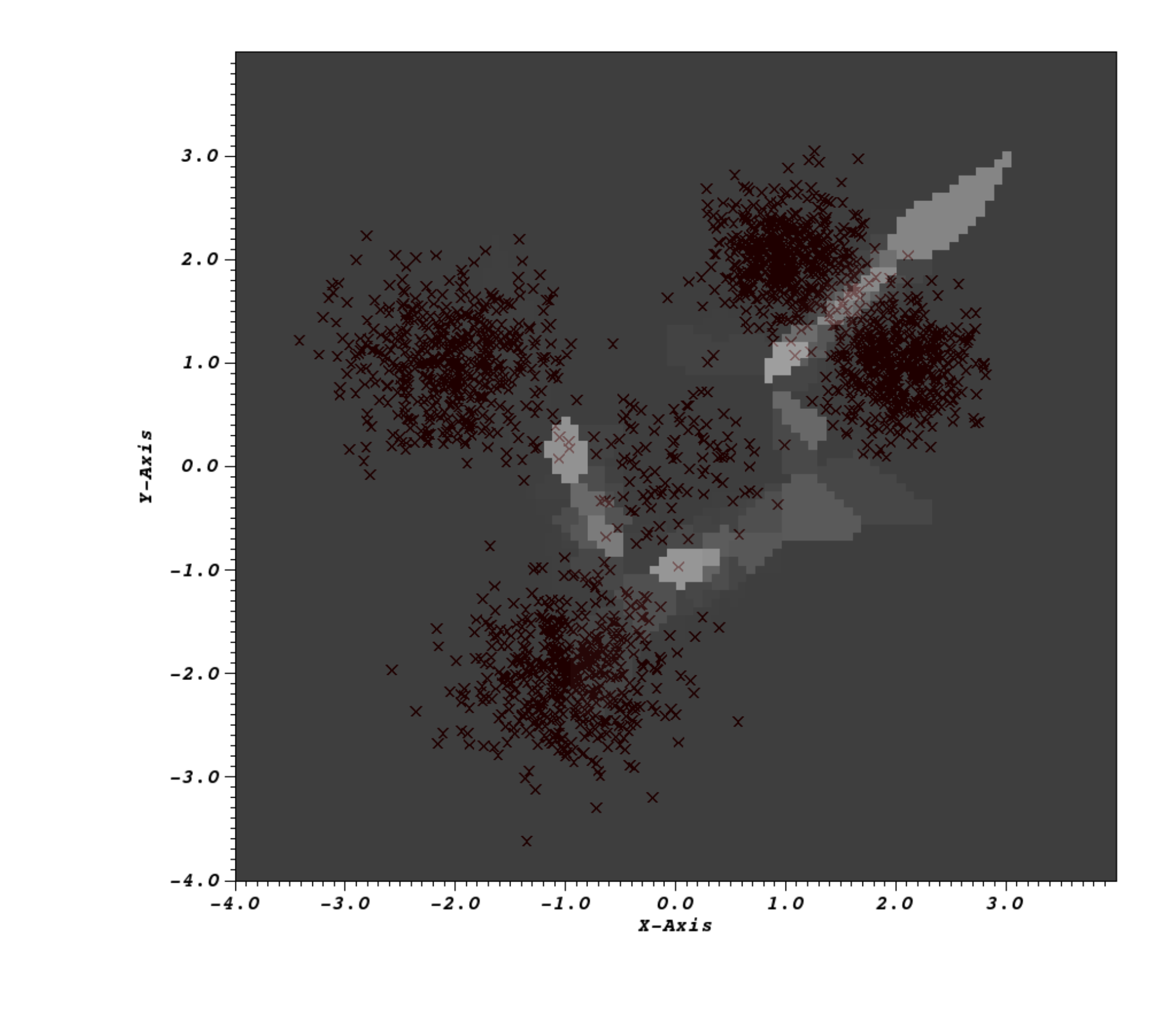}
  \caption{Heat map}
  \label{fig:viz3}
\end{figure}

\subsection{Extensions}
\label{sec:extensions}

Our definition of affinity is not limited to Euclidean spaces. It can be generalized to a variety of spaces merely by modifying the way in which we construct the Voronoi diagrams. In all cases, the resulting affinity scores will result from a volume computation over polyhedra. 

\paragraph{Giving clusters varying importance: density-based methods}
\label{sec:weighted-regions}

The definition of affinity assumes that all clusters are equally important, since the Voronoi-based construction treats distances from all centers the same way. However, cluster size is an important factor when considering whether a point has been correctly labeled. 
Intuitively, dense clusters have a larger region of influence and the affinity scores should account for this. Consider therefore a generalized clustering instance where each cluster $C_i$ has an associated weight $w_i$, with a larger $w_i$ indicating greater importance. 
Instead of constructing the Voronoi diagram, we will construct the \emph{power diagram} defined in Section~\ref{sec:preliminaries}. Specifically, the region of influence $R_i$ for cluster $C_i$ will be defined as the set $R(C_i) = \{ x | d^2(p_i, x) - w_i \le d^2(p_j, x) - w_j\}$. We compute the affinity vector as before, with the weight of $x$ set appropriately depending on the weight function used. For example, if $w(C_i) = |C_i|/n$, then $w(x) = 1/n$.

\begin{figure}[htbp]
  \centering
  \subfigure[A data set with 100 points in each cluster]{\includegraphics[width=3in]{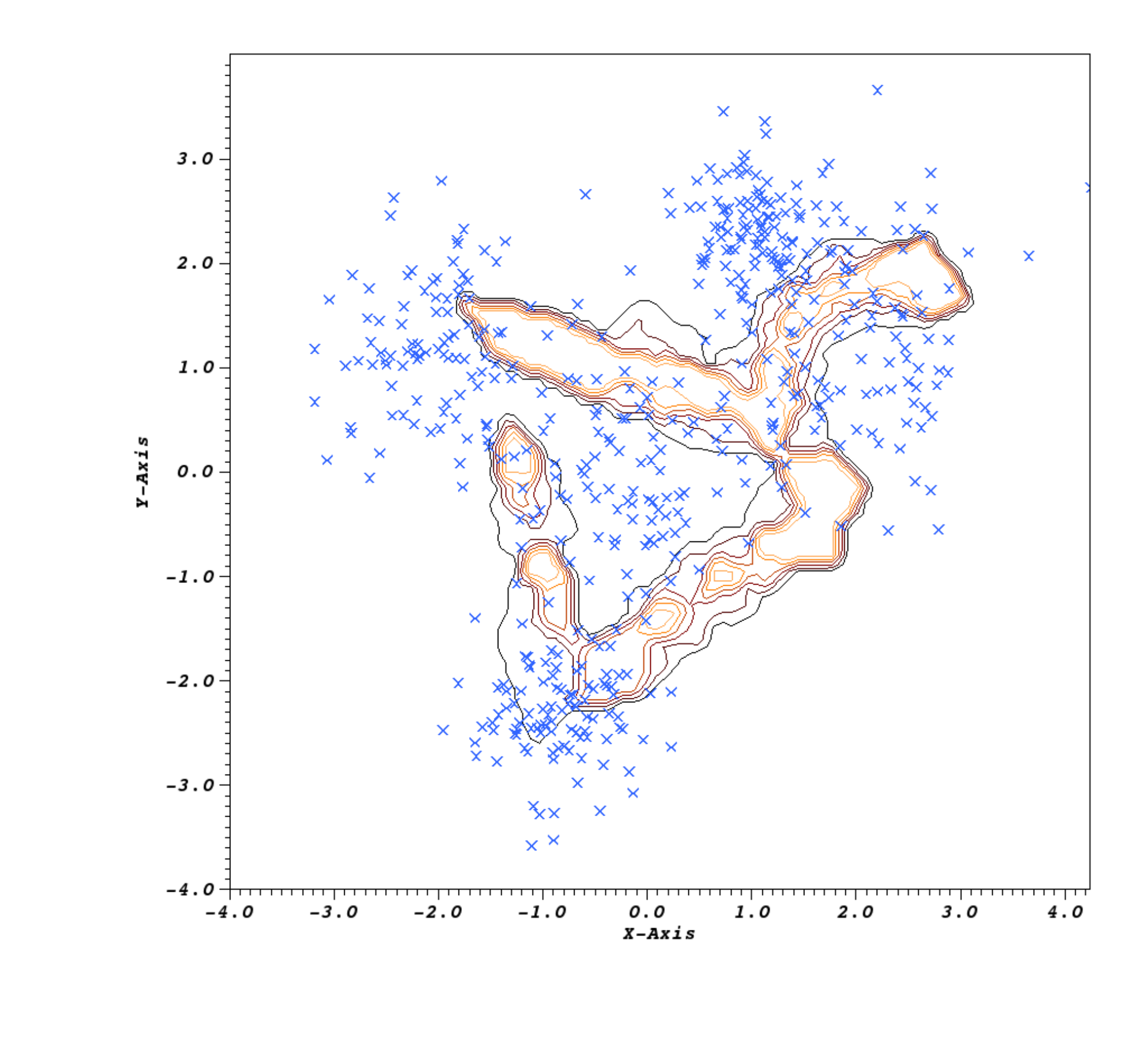}}
  \subfigure[A data set 500 points in each of four cluster, and 100 in the center.]{\includegraphics[width=3in]{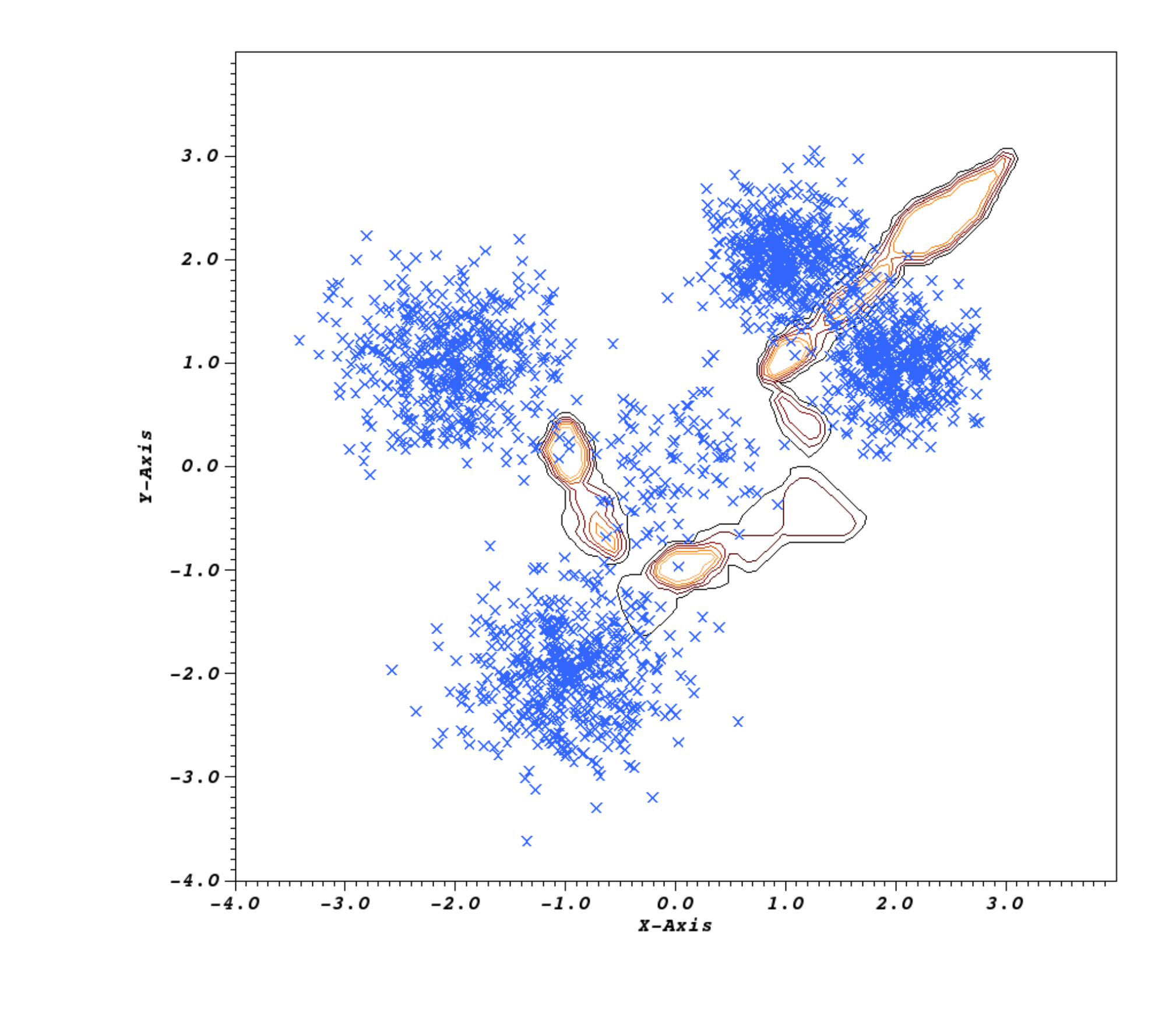}}
  \caption{\label{fig:density}}
\end{figure}

Consider the examples depicted in Figure~\ref{fig:density}. The lefthand figure has 100 points in each of five clusters, and the right-hand figure has 500 points in each of four outer clusters and 100 points in the center cluster. Notice that there is a lot more instability (as seen by the contours) in the sparser example, much of which is due to the presence of the central cluster. However, once the density of the outer clusters increases, the effect of the inner cluster is much weaker, and there are fewer unstable regions. 

\paragraph{Bregman Divergences}
\label{sec:bregman-divergences}
While the Euclidean distance is a common underlying metric for clustering, the Bregman divergences\cite{bregman} are also very important. Clustering with Bregman divergences is equivalent\cite{banerjee} to mixture estimation for a corresponding family of distributions. Mixture estimation with Gaussians leads to the use of the Euclidean distance (squared), the multinomial distributions give rise to the Kullback-Leibler distance, and so on. 

Let $D_\phi(p \mid q) = \phi(p) - \phi(q) - \langle \nabla \phi(q), p - q\rangle$ be the Bregman divergence from $p$ to $q$. It has been shown\cite{bregvor} that the \emph{Bregman Voronoi} diagram on a collection of sites $s_1, \ldots, s_k$ defined as $V_i = \{ p | D_\phi(p, s_i) \le D_\phi(p, s_j) \forall j \ne i\}$ is a convex polyhedron just like in the case of the Euclidean distance\footnote{An easy way to see this is to solve the equation $D_\phi(p \mid s_i) = D_\phi(p \mid s_j)$ for the bisector.}. Thus, defining the region of influence $R_i$ to be $V_i$, we can define affinity scores appropriately. Note that the above procedure for incorporating weights on clusters also generalizes, by defining $R(C_i) = \{ x | D_\phi(x \mid p_i) - w_i \le D_\phi(x \mid p_j) - w_j\}$. 

\paragraph{Kernels}
\label{sec:kernels}

If we have a distance space defined by a kernel function\cite{kernels}, we can define regions of influence in a similar manner. The $n\times n$ Gram matrix defined by the kernel yields an $n$-dimensional Euclidean space (via the ``kernel trick''), and then the problem reduces to computing Voronoi regions and estimating volumes in this space. An explicit embedding into $n$-dimensional space is still prohibitively expensive, and so we will explain how to address the problem using approximate lifting maps. 

\section{Estimating Affinity}

The many different ways of defining affinity scores via regions of influence all reduce to the following: given a set of representatives $C = \{c_1, \ldots, c_k\}$ and a query point $x$, estimate the volume of a single cell in the Voronoi diagram of $C$ or $C \cup \{x\}$, and estimate the volume of the intersection of two such cells. 

In two and three dimensions, it is relatively easy to compute these quantities efficiently. In two dimensions, the Voronoi (or weighted Voronoi) diagram of $k$ points can be computed in time $O(k \log k)$\cite{cgbook}, and the intersection of two convex polygons can be computed in $O(k)$ time\cite{toussaint}. Any polygon with $k$ vertices can be triangulated on $O(k)$ time using $O(k)$ triangles, and then the area can be computed in $O(k)$ time ($O(1)$ time per triangle). In three dimensions, computing the Voronoi diagram takes $O(k^2)$ time. This is the dominant term in the running time, as computing the intersection of two convex polyhedra can be done in linear time \cite{chazelle}. Tetrahedralizing the convex polyhedron can also be done efficiently\cite{lennes1911theorems}. 

However, this direct approach to volume computation does not scale. In general, a single cell in the Voronoi diagram of $k$ points in $\reals^d$ can have complexity $O(k^{\lceil d/2\rceil})$. We now propose an alternate strategy that provably approximates the affinity scores to any desired degree of accuracy using random sampling. 

Let $V_x$ be the Voronoi cell of $x$ in the Voronoi diagram of $C \cup \{x\}$. We say that $y$ is \emph{stolen from} $s(y) = c_i$ if (i) $y \in V_x$ and (ii) $y$'s second nearest neighbor is $c_i$. We can then write $\alpha_i = \frac{ \vol(\{ x \mid s(x) = c_i \}) }{\vol(V_x)}$. 
Note that given a point $x$ and any point $y$, we can verify in $O(k)$ time whether $y \in V_x$ and also compute $s(y)$ by direct calculation of the appropriate distance measure. 

Let $(\alpha_1, \alpha_2, \ldots, \alpha_k)$ be the affinity scores for $x$. Suppose we now sample a point $y$ uniformly at random from $V_x$. We can find $s(y)$ in $O(k)$ time and this provides one update to $\alpha_i$. The number of such samples needed to get an accurate estimate of each $\alpha_i$ is given by the theory of $\eps$-samples. Let $\mu$ be a measure defined over $X$ and let $\mathcal{R}$ be a collection of subsets of $X$. An \emph{$\eps$-sample} with respect to $(X, \mathcal{R})$ and $\mu$ is a subset $S \subset X$ such that for any subset $R \in  \mathcal{R}$, 
  \[ \Bigl|\frac{\mu(S \cap R)}{\mu(S)} - \frac{\mu(R)}{\mu(X)}\Bigr| \le \eps. \]
By standard results in VC-dimension theory\cite{geometric-approx}, a random subset of size $O(d/\eps^2 \log 1/\eps)$ is an $\eps$-sample for a range space $(X, \mathcal{R})$ of VC-dimension\cite{geometric-approx} $d$.

If we now consider the discrete space $[1\ldots k]$ with the measure $\mu(i) = \alpha_i$, then the set of ranges $\mathcal{R}$ is the set of singleton queries $\{1 \ldots k\}$, and the VC-dimension of $([1\ldots k], \mathcal{R})$ is a constant. This means that if we sample a set $S$ of $O(d/\eps^2 \log 1/\eps)$ points from $V_x$, and set $\tilde{\alpha}_i  = \frac{|\{x \in S \mid s(x) = i\}}{|S|}$, then 
$|\tilde{\alpha}_i - \alpha_i| \le \epsilon$ for all $i$. 

\subsection{Sampling from $V_x$}
\label{sec:sampling-from-v_x}

We now have a strategy to estimate the affinity scores of $x$. Sample the number of points from $V_x$ as prescribed above and then estimate $\tilde{\alpha}_i$ by computing the owners of samples. A simple sampling strategy would be as follows. Find a ball that encloses $V_x$. Sample uniformly from within this ball, and then reject any points outside $V_x$. Unfortunately, as the dimension of the space increases, the number of rejections grows exponentially with $d$, and so the time required to produce even one sample is exponential in $d$. For example, our experiments show that in twenty dimensions, over one thousand points are rejected for each good sample. 

\begin{figure}[htbp]
  \centering
  \includegraphics[width=4in]{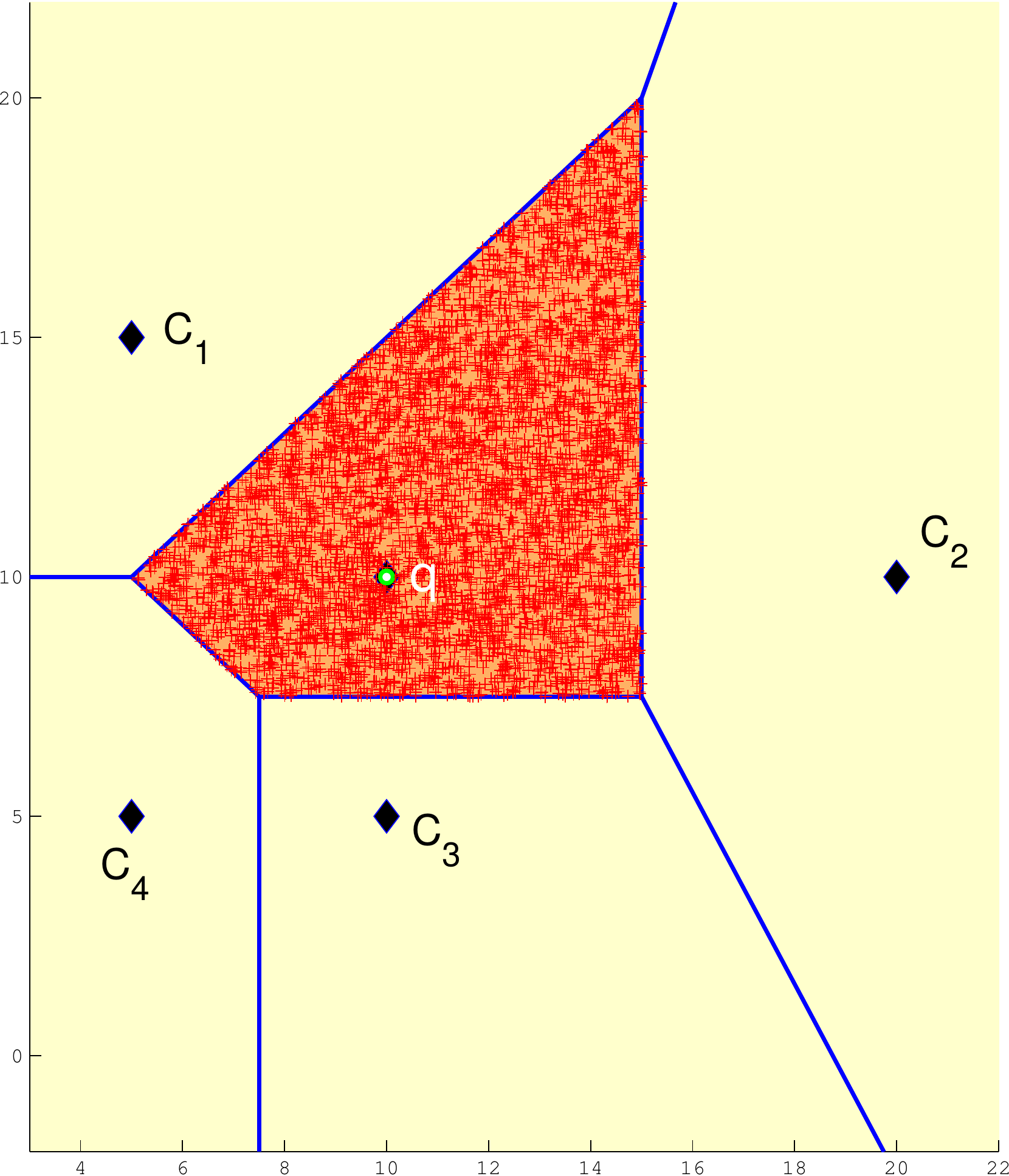}
  \caption{Illustration of \hitandrun for sampling from a Voronoi cell. Samples are shown in red. }
  \label{fig:hitandrun}
\end{figure}

The problem of sampling from a convex polytope in high dimensions has been studied extensively. The main focus of these efforts has been to estimate the volume of a convex polytope via sampling, following the groundbreaking randomized polynomial time algorithm of Dyer, Frieze and Kannan. At a high level, these are all MCMC methods: they use different random walks  to extract a single uniform sample from the polytope. One of the most effective strategies in practice for doing this is known as \emph{hit-and-run}\cite{smith}. It works as follows. Starting with  some point $x$ in the desired polytope $K$, we pick a direction at random, and then pick a point uniformly on the line segment emanating from $x$ in that direction and ending in the boundary of $K$. We refer to this step as \hitandrun. It has been shown\cite{lovasz-hr} that this random walk mixes very well, making $O(d^3)$ calls to a membership oracle to produce a single sample (under some technical assumptions). Figure~\ref{fig:hitandrun} illustrates 
the distribution of samples using \hitandrun for the Voronoi cell of the point $q$. 

Algorithm~\ref{alg:nni} (\affinityalgo) summarizes the process for computing the affinity score of a single point.

\begin{algorithm}[t]
\caption{\SamplePolytope\label{alg:polytope}}
\textbf{Input:} Collection of halfplanes $\mathcal{H}$ defining convex region $K = \cap_{h \in \mathcal{H}} h$, number of samples $m$.

\textbf{Output:} $m$ points uniformly sampled from $K$.\raggedright
  \begin{algorithmic}
  \STATE Construct affine transform $T$ such that $TK$ is centered and isotropic.
  \STATE Fix \emph{burn-in} parameter $b$
  \STATE Run \hitandrun for $d$ steps on $TK$, ending in $z = z_0$
  \FOR{$i = 1 \ldots m$}
  \STATE Set $z_i$ to be result of one \hitandrun move from $z_{i-1}$
  \ENDFOR
  \STATE Return $(T^{-1}z_1, \ldots, T^{-1}z_m)$. 
  \end{algorithmic}
\end{algorithm}
\begin{algorithm}
  \caption{\affinityalgo: Computing the affinity score for a point\label{alg:nni}}
\textbf{Input:} A clustering $\mathcal{C} = C_1, C_2, \ldots, C_k$ with representatives $c_1, \ldots, c_k$ and a point $x$.

\textbf{Output:} Affinity vector $(\alpha_1, \ldots, \alpha_k)$ for $x$\raggedright
\begin{algorithmic}
  \STATE $m \leftarrow \frac{c}{\eps^2}\log \frac{1}{\eps}$
  \STATE Set all $\alpha_i \leftarrow 0$
  \FOR{$j = 1 \ldots k$}
  \STATE Set $\mathcal{H}_j$ as the halfplane supporting $V_x$ with respect to $c_j$ in the Voronoi diagram.
  \ENDFOR
  \STATE Call \SamplePolytope$(\{ \mathcal{H}_1, \ldots, \mathcal{H}_k \}, m)$ to generate $m$ samples $z_1, z_2, \ldots z_m \in V_x = \cap \mathcal{H}_j$. 
  \FOR{$i = 1 \ldots m$}
  \STATE Compute $s = \arg\min_{j=1\ldots k} d(z_i, c_j)$. 
  \STATE $\alpha_s = \alpha_s + 1/m$
  \ENDFOR 
  \STATE Return $(\alpha_1, \ldots, \alpha_k)$. 
\end{algorithmic}

\end{algorithm}

\paragraph{Kernel spaces}
In Section~\ref{sec:extensions} we described how to extend the notion of affinity vectors to kernel spaces by using the representer theorem and the Gram matrix representation of the kernel values. Unfortunately, the resulting space has $n$ dimensions, which makes \affinityalgo very expensive. A better approach is to use an approximate representation of the kernel using a lifting map like hash kernels\cite{shi2009hash} or the random Fourier method\cite{Rahimi2007}. It has been shown that the number of dimensions required to yield an $\epsilon$-approximation to the true kernel distances is roughly $O(\log n)/\eps^2$, which is significantly better than $n$ for large enough $n$. 

\paragraph{Reducing dimensionality}
The above sampling procedure runs in time $O(d^3)$ per point. However, $d$ can be quite large. We make one final observation that replaces terms involving $d$ by terms involving $k \ll d$ for Euclidean distance measures (or Euclidean distances derived from a kernel).

The Voronoi diagram of $k$ points in $d$ dimensions, where $k < d$, has a special structure. The $k$ points together define a $k-1$-dimensional subspace $\mathcal{H}$ of $\reals^d$. This means that any vector $p \in \reals^d$ can be written as $p = u + w$ where $u \in \mathcal{H}$ and $w \perp u$. The Euclidean distance $\|p - p'\|^2$ can be written as $\|u - u'\|^2 + \|w - w'\|^2$. In particular, this means that in any subspace of the form $\mathcal{H} + w$ for a fixed $w \perp \mathcal{H}$, the distance between two points is merely their distance in $\mathcal{H}$. 

Therefore, each Voronoi cell $V$ can be written as $V' + \mathcal{H}^\perp$, where $V' \subset \mathcal{H}$ and $\mathcal{H}^\perp$ is the orthogonal complement of $\mathcal{H}$ consisting of all vectors orthogonal to $\mathcal{H}$. Thus, we can project all points onto $\mathcal{H}$ while retaining the same volume ratios as in the original space. This effectively reduces the problem to a $k$-dimensional space. The actual projection is performed by doing an singular value decomposition on the $k \times d$ matrix of the cluster representatives. Once this transformation is done, we call \affinityalgo as before. 

The resulting algorithm computes all affinity scores in time  $O(nk^3\log(1/\eps)/\eps^2)$.

\section{Experiments}

In two and three dimensions, affinity scores can be calculated via direct volume computations. We use built-in routines provided by \textsc{CGAL}(\url{http://www.cgal.org}) to compute the scores exactly and validate our sampling-based algorithm. For higher dimensional data, we perform the initial data transformation (if needed) in \texttt{C} and use a native routine for \hitandrun in \textsc{Matlab}. All experiments are run on a Intel Quad Core CPU 2.66GHz mahine with 4GB RAM. Reported times represent the results of averaging over 10 runs. 

We created two synthetic datasets in $\reals^2$ namely, \textit{2D5C-500} for which data is drawn from $5$ Gaussians to produce $5$ visibly separate clusters with $100$ points each and \textit{2D5C-2100} which adds $400$ points each to the 4 clusters in the corners. We also use 2 different datasets from the UCI repository (Iris, Soybean) along with the test set of MNIST (\url{http://yann.lecun.com/exdb/mnist/}) handwritten digits that contains 10,000 examples respectively in  $\reals^{784}$. 

\subsection{Case Studies}
\label{sec:case-studies}

We start with three case studies of applications that show a demonstrable benefit from knowing affinity scores for points. In all cases, our goal is to emphasize the flexibility and usability of affinity scores, rather than claim that affinity scores can solve all the problems listed here. 

\paragraph{Active Clustering}
\label{sec:active-clustering}

The idea of active clustering is to select a few points that are most likely to influence the formation of good clusters. We show how affinity scores can be used to guide the search for such points, reducing the number of points that the clustering algorithm needs to process.  

We run a $k$-means++\cite{kmeansplus} seeding to initialize $k$ cluster centers. We then compute affinity scores for all points. We fix a fraction $0 < \alpha < 1$ (set by cross validation) and then select a sample of points of size $2 \alpha\sqrt{n}$ from the pool of stable points, selecting the remaining $2 (1-\alpha)\sqrt{n}$ points at random from the unstable pool. We then cluster this small set using $k$-means and assign all remaining points to their nearest cluster center. We compare this to doing a straight $k$-means implementation. As we can see in Table~\ref{tbl:performance-active}, even though the data size is reduced considerably (see Table~\ref{tbl:data-active}), the quality of the solutions remains the same. 

The memory footprint of this approach is considerably smaller. A standard implementation of $k$-means needs an $n \times n$ matrix of distances of $\Theta(n^2)$ space, which in our procedure is reduced to $O(n)$ space. While the $k$-means algorithm can be implemented without explicitly materializing the distances, it then must pay a price in time to recompute each distance, which for large-dimensional data would be prohibitive. 

\begin{table}[htbp]
\small
\begin{center}
  \begin{tabular}{ | l | l | l | l | l |}
    \hline
    Dataset & Points & Samples & $\#$ Stable & $\#$ Unstable  \\ \hline \hline
    \textit{2D5C-500} & 500 & 50 & 30 & 20  \\ \hline
    MNIST test data & 10000 & 200 & 160 & 40 \\ \hline
  \end{tabular}
\caption{\sffamily Data setup for Active Clustering.\label{tbl:data-active}}
\end{center}
\end{table}

\begin{table*}[t]
\small
\begin{center}
  \begin{tabular}{ | l || l | l || l | l || l | l || l | l |}
    \hline
    & \multicolumn{2}{|c||}{\liftemd} & \multicolumn{2}{|c||}{Rand Distance} & \multicolumn{2}{|c||}{$\%$ Unstable points}\\
    \cline{2-7}
    Dataset & d(\textit{$P_{AC}$},\textit{$P_{REF}$}) & d(\textit{$P_{KM}$},\textit{$P_{REF}$}) & d(\textit{$P_{AC}$},\textit{$P_{REF}$}) & d(\textit{$P_{KM}$},\textit{$P_{REF}$}) & $P_{AC}$ & $P_{KM}$\\ \hline \hline
    \textit{2D5C-500} & 0.16 & 0.18 & 0.22 & 0.26 & 20 & 18 \\ \hline
    MNIST test data & 0.14 & 0.13 & 0.12 & 0.10 & 14 & 17 \\ \hline
  \end{tabular}
\caption{\sffamily Comparing partitions generated by active clustering ($P_{AC}$) and $k$-means ($P_{KM}$) w.r.t. reference Partition ($P_{REF}$). Smaller numbers indicate better performance.\label{tbl:performance-active}}                                                                                                                                                            
\end{center}
\end{table*}

\paragraph{Validating Consensus Clustering}
\label{sec:consensus-clustering}

When we compute a consensus clustering from a collection of clusterings, it is hard to determine whether the new clustering is in any sense better due to calibration issues on the quality score.  The ``number of stable points'' is a different measure that can be used to evaluate the quality of the consensus. Intuitively, the consensus is improved if points have moved from being unstable to being stable. 

We take four data sets for which we have a reference clustering and run five clustering algorithms on each. We then run a spatially-aware consensus procedure\cite{DBLP:conf/sdm/RamanPV11} and obtain a consensus clustering of each data set. We now compare this consensus clustering to one of the base clusterings and report two distances (\liftemd\cite{DBLP:conf/sdm/RamanPV11} and the Rand Distance) as well as the change in the number of unstable points. The results are tabulated in Table~\ref{tbl:performance-consensus}. 

\begin{table*}[t]
\small
\begin{center}
  \begin{tabular}{ | l || l | l || l | l || l | l || l | l |}
    \hline
    & \multicolumn{2}{|c||}{\liftemd} & \multicolumn{2}{|c||}{Rand Distance} & \multicolumn{2}{|c||}{$\%$ Unstable points}\\
    \cline{2-7}
    Dataset & d(\textit{$P_{CON}$},\textit{$P_{REF}$}) & d(\textit{$P_{KM}$},\textit{$P_{REF}$}) & d(\textit{$P_{CON}$},\textit{$P_{REF}$}) & d(\textit{$P_{KM}$},\textit{$P_{REF}$}) & $P_{CON}$ & $P_{KM}$\\ \hline \hline
    \textit{2D5C-500} & 0.15 & 0.18 & 0.20 & 0.26 & 10 & 18 \\ \hline
    IRIS & 0.10 & 0.12 & 0.11 & 0.15 & 8 & 8 \\ \hline
    Soybean & 0.27 & 0.32 & 0.15 & 0.19 & 22 & 30 \\ \hline
    MNIST test data & 0.10 & 0.13 & 0.09 & 0.10 & 15 & 17 \\ \hline
  \end{tabular}
\caption{\sffamily Using affinity scores to evaluate performance of consensus.\label{tbl:performance-consensus}}                                  
\end{center}
\end{table*}

In all data sets, the consensus clustering is closer to the reference clustering than the base clustering that uses $k$-means. This is of course why we compute consensus in the first place. But looking closer, we see that for 2D5C-500 and Soybean, the fraction of unstable points reduces considerably indicating a tighter collection of clusters. In contrast, the reduction for MNIST and IRIS is small. Note that in all cases, the distance reductions are the same and do not distinguish the sets. This shows two things: firstly, stability correlates well with other measures of clustering distance. Secondly, the number of unstable points allows us to understand consensus quality more deeply than just via a global distance measure. 

\paragraph{Incremental Clustering}
\label{sec:incr-clust}

As a final demonstration of the versatility of affinity scores, consider the problem of incremental clustering. The typical approach to incremental clustering \cite{birch,DBLP:reference/db/Venkatasubramanian09} is to  cluster a block of data, build a reduced representation of the cluster, and then read the next block of data and update this representation. We can use stable points to define the cluster and its centroid, while keeping unstable points around unassigned pending further information. Each time a new batch of data comes in, we determine the stable points in this batch with respect to the current clustering, update the centers, and then check if any more points have become stable or unstable. The process then repeats. Note that we need only maintain cluster centroids for high quality clusters and know which points to defer decisions on. Table~\ref{tbl:performance-incremental} illustrates the results of this on the MNIST test data, where at each stage we compare the clustering obtained to the reference clustering. We see that the clustering produced is of quite high quality inspite of the severe size reduction (10 cluster centers and a decreasing number of unstable points).  

\begin{table}[htbp]
\small
\begin{center}
  \begin{tabular}{ | l | l | l | l |}
    \hline
    Dataset & $\# Points$ & \liftemd & $\%$ Unstable points \\ \hline \hline
    MNIST batch 1 & 2000 & 0.16 & 20  \\ \hline
    MNIST batch 2 & 3000 & 0.14 & 18 \\ \hline
    MNIST batch 3 & 1000 & 0.15 & 18  \\ \hline
    MNIST batch 4 & 2500 & 0.14 & 16 \\ \hline
    MNIST batch 5 & 1500 & 0.13 & 16 \\ \hline   
  \end{tabular}
\caption{\sffamily Running incremental clustering on MNIST test data.\label{tbl:performance-incremental}}
\end{center}
\end{table}

\subsection{Validation}
\label{sec:validation}

\subsubsection{Concept Validation}
\label{sec:concept-validation}

The premise of the affinity scores is that low affinity corresponds to points that are difficult to cluster. One way to test this idea directly is to evaluate the affinity scores for data that admits a verified labeling. The MNIST database (\url{http://yann.lecun.com/exdb/mnist/}) of digits is useful to test this because we can visually inspect the digits with low affinity to see whether they have ambiguous labels. 

\begin{figure}[htbp]
  \centering
  \includegraphics[width=2in]{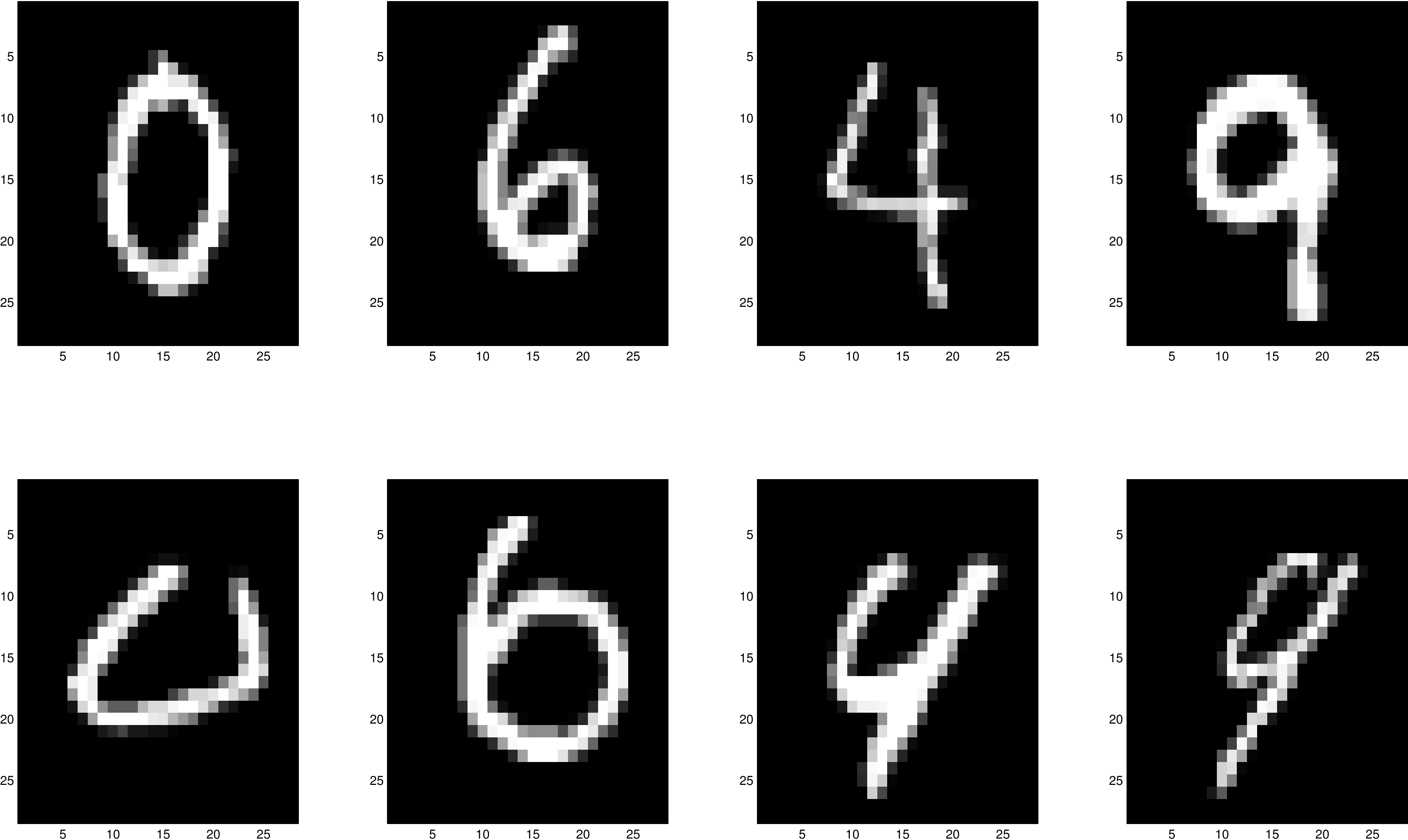}
  \caption{Results of running $k$-means on MNIST training data. First
    row: high affinity. (L-R) 0.96, 1.0, 1.0,
    0.92. Second row: low affinity: (L-R) 0.38, 0.46,
    0.34, 0.42.}
  \label{fig:mnist-sensitivity}
\end{figure}

We run a $k$-means algorithm on the MNIST test data and compute affinity scores of the points. We then visualize one example each from the stable and unstable regions. Figure~\ref{fig:mnist-sensitivity} shows the results. The first row shows points that had high affinity in the clustering (close to 1.0 in each case). We can see that the digits are unambiguous. The second row shows digits from the unstable region (affinity scores less than 0.5). Notice that in this case the digits are far more blurred. In fact, the $4$ and $9$ look similar, as do the $0$ and $6$. This indicates that low affinity does in fact correlate with ambiguous labels as indicated. 

\subsubsection{Score Validation}
\label{sec:score-validation}

We can validate the scores produced by sampling against the exact scores in two and three dimensions using exact volume computations. Table~\ref{tbl:performance-time-approx} illustrates this for the 2D5C and 3D5C data sets. We note that these error reports come from choosing $1000$ samples after a burn-in of $1000$ samples (this corresponds to an error $\eps = 0.04$). As we can see, the reported error is well within the predicted range.

\subsection{Running times}
\label{sec:running-times}

Table~\ref{tbl:performance-time-approx} also presents running times for the affinity score computation. We note that the running times reported are the total for computing the affinity scores for \emph{all} points. We only report the time taken by the sampler; the preprocessing affine transformation is dominated by the sampling time. In all cases, we used 1000 samples to generate the estimates.  Note that the procedure is extremely fast, even for the very high dimensional MNIST data. 

\begin{table*}[htbp]
\small
\begin{center}
  \begin{tabular}{ | l | l | l | l | l | l |}
    \hline
    Dataset & $\#$Points & $\#$Dimensions & $\#$Clusters & Runtime (sec) & Empirical Approximation \\ \hline \hline
    \textit{2D5C-500} & 500 & 2 & 5 & 0.11 $\pm$ 0.005 & $\pm$ 0.02 \\ \hline
    \textit{3D5C-500} & 500 & 3 & 5 & 0.19 $\pm$ 0.008 & $\pm$ 0.035 \\ \hline
    IRIS & 150 & 4 & 3 & 0.24 $\pm$ 0.012 & - \\ \hline
    Soybean & 47& 35 & 4 & 0.31 $\pm$ 0.08 & - \\ \hline
    MNIST test data & 10000 & 784 & 10 & 0.58 $\pm$ 0.5 & - \\ \hline
  \end{tabular}
\caption{\sffamily Runtimes and empirical approximation to exact affinity: Estimating affinity for one point by generating 1000 samples.\label{tbl:performance-time-approx}}
\end{center}
\end{table*}

\section{Conclusion}

In this paper we present a method to validate the assignment of points to clusters in a clustering. We show different ways in which identifying points that are ``unstable'' can enhance or illuminate downstream clustering tasks, and validate the notion of a local affinity score against clusterings where the ground truth is known. 

We view this work as part of a larger effort to personalize validation mechanisms in data mining. In future work we plan on incorporating ideas from topological data mining to add more dimensions to the validation. We hope to develop better visualizations to accompany this method. And more generally we plan on studying other unsupervised learning tasks where local validation is important. 

\bibliographystyle{abbrv}
\bibliography{refs}  
\end{document}